\documentclass[letterpaper, 10 pt, conference]{ieeeconf}  
\usepackage{times}

\usepackage{multicol}
\usepackage[bookmarks=true]{hyperref}
\usepackage{amsmath}
\usepackage{amsfonts}
\usepackage{xcolor} 
\usepackage{graphicx}
\usepackage{algorithm}
\usepackage{algpseudocode}
\usepackage{diagbox}
\usepackage{caption}
\usepackage{subcaption}
\usepackage{bbm}
\usepackage{color, soul} 
\newcolumntype{?}{!{\vrule width 1pt}}

\IEEEoverridecommandlockouts                              

\title{\LARGE \bf
Reinforcement Learning in a Safety-Embedded MDP with Trajectory Optimization
}

\author{Fan Yang$^{1}$, Wenxuan Zhou$^{1}$, Zuxin Liu$^{1}$, Ding Zhao$^{1}$, David Held$^{1}$
\thanks{$^{1}$All authors are with Carnegie Mellon University, Pittsburgh, PA 15213, USA.  {\tt\small fanyangr@umich.edu}
        }%
\thanks{This material is based upon work supported by the National Science Foundation under Grant No. IIS-1849154 and the United States Air Force and DARPA under Contract No. FA8750-18-C-0092.}
}
\begin{document}
\maketitle
\thispagestyle{empty}
\pagestyle{empty}
\begin{abstract}
Safe Reinforcement Learning (RL) plays an important role in applying RL algorithms to safety-critical real-world applications, addressing the trade-off between maximizing rewards and adhering to safety constraints. This work introduces a novel approach that combines RL with trajectory optimization to manage this trade-off effectively. Our approach embeds safety constraints within the action space of a modified Markov Decision Process (MDP). The RL agent produces a sequence of actions that are transformed into safe trajectories by a trajectory optimizer, thereby effectively ensuring safety and increasing training stability. 
This novel approach excels in its performance on challenging Safety Gym tasks, achieving significantly higher rewards and near-zero safety violations during inference. The method's real-world applicability is demonstrated through a safe and effective deployment in a real robot task of box-pushing around obstacles. Further insights are available from the videos and appendix on our website: \href{https://sites.google.com/view/safemdp}{https://sites.google.com/view/safemdp}.
\end{abstract}
\section{INTRODUCTION}
\label{sec:intro}
Reinforcement Learning (RL) has seen tremendous success in solving sequential decision-making problems \cite{mnih2013playing, akkaya2019solving, andrychowicz2020learning, kiran2021deep, pierson2017deep, levine2016end}. However, deploying these algorithms in real-world robotic systems raises safety concerns, particularly in safety-critical applications like obstacle avoidance, autonomous driving, and human-robot interactions. 
A common approach to tackle safety in RL is to define the task under the Constrained Markov Decision Process (CMDP) framework, which defines a constrained optimization in which the agent must maximize the reward while satisfying safety constraints~\cite{altman1999constrained, achiam2017constrained, ray2019benchmarking, as2022constrained}. Unfortunately, most constrained optimization-based methods struggle with the delicate balance between reward maximization and constraint satisfaction during the learning process \cite{ray2019benchmarking, liu2022constrained}, often leading to unstable training. Underestimates of safety cost values can lead to the convergence of unsafe policies. Conversely, overestimates of the safety cost values may result in conservative exploration and suboptimal task performance. 

\begin{figure}
    \centering
    \includegraphics[width=0.48\textwidth]{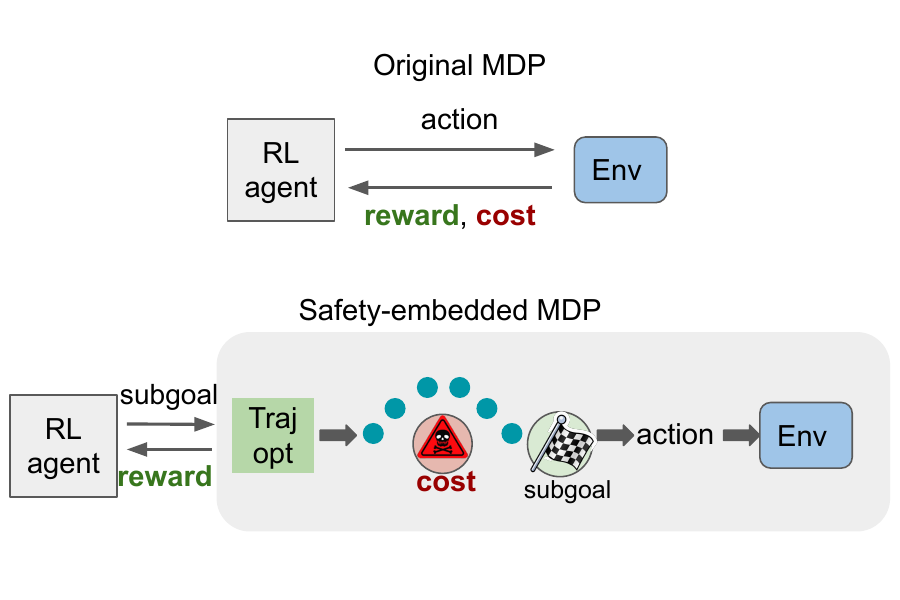}
    \caption{Compared to previous methods, in which the RL agent optimizes the reward and safety constraints simultaneously (left), our method operates in a modified MDP (right). The modified MDP is embedded with a trajectory optimizer to ensure constraint satisfaction. The RL agent outputs a subgoal for the safe trajectory optimizer and hence the RL agent only needs to optimize explicitly for the reward, leading to much better performance with fewer safety violations.}
    \label{fig:pull_fig}
\end{figure}

To address this challenge, we propose a novel approach that incorporates trajectory optimization within an RL framework, providing a powerful tool for handling safety constraints defined around obstacle avoidance. 
The RL agent operates in a modified MDP, embedded with a trajectory optimization algorithm to ensure safety. 
Specifically, the RL agent outputs actions in a high-level action space, which are transformed into low-level actions via a trajectory optimizer which is restricted to generating safe trajectories.  The trajectory optimizer is treated as part of the transition dynamics of the modified MDP.
This approach allows the RL agent to optimize an unconstrained objective in the modified MDP, leading to faster and more stable training, improved performance, and better safety constraint satisfaction.  The framework of our method is shown in Fig.~\ref{fig:pull_fig}.

We demonstrate the efficacy of our method through comprehensive experimentation, greatly outperforming previous approaches in terms of both reward maximization and safety constraint satisfaction in complex contact-rich simulated and real-world settings. We focus on the very challenging block-pushing task from Safety Gym~\cite{ray2019benchmarking}, in which the goal is to train an agent to push a block to a goal while avoiding obstacles. Previous safe RL methods have failed to make reasonable progress on this task; this work represents the first safe RL method to achieve a high level of task success and constraint satisfaction at this difficult task which involves both long-horizon reasoning as well as reasoning about both contacts and obstacle avoidance.

Notably, our method achieves over 50\% success rate in challenging PointPush2 and CarPush2 tasks, which is about 20 times higher than the strongest baselines with a similar safety performance, while maintaining very low levels of cost violation. 
\section{RELATED WORK}
\label{sec:related_work}
\textbf{Safe Reinforcement Learning:}
Safe reinforcement learning (RL) methods can be generally categorized into model-based and constrained optimization approaches. The former, including methods proposed by Pham et al. \cite{pham2018optlayer} and Dalal et al. \cite{dalal2018safe}, employ a combination of model-free methods with safety checks to achieve constrained exploration.
Other methods reframe safety constraints by shielding functions to monitor and correct policy action output ~\cite{ jayant2022model, alshiekh2018safe}. 
However, these often 
require substantial domain knowledge which creates difficulties in scaling with the number of tasks that we want robots to perform~\cite{berkenkamp2017safe,koller2018learning}.

Another approach is to use constrained optimization methods like primal-dual approaches~\cite{ding2020natural, bohez2019value, ray2019benchmarking, 
as2022constrained} alternating between policy parameter optimization and dual variable updating. Despite their intuitiveness, these techniques suffer from training instabilities~\cite{chow2018lyapunov, xu2021crpo}. Attempts to improve these, such as introducing a KL-regularized policy improvement mechanism, often lead to high sample complexity or poor constraint satisfaction performance~\cite{achiam2017constrained, yang2020projection, liu2022constrained}. Our method attempts to address these challenges by utilizing trajectory optimization in the modified MDP to enhance performance and safety during both training and deployment.

\textbf{Trajectory Optimization:}
In our method, the trajectory optimizer is used to solve a path-planning problem with obstacle avoidance. Path planning with obstacle avoidance is a canonical problem in motion planning~\cite{likhachev2003ara,koenig2002d, fiorini1998motion}. 

When it comes to object interactions, such as pushing a box, the nonlinear dynamics involved in such interactions make trajectory optimization more challenging, although
some trajectory optimization methods based on Model Predictive Control (MPC) can be applied to contact-rich tasks as well~\cite{howell2022predictive, howell2022calipso, howell2022trajectory}. However, when it comes to long-horizon tasks, significant time and computing resources are usually needed to compute effective solutions. 

\textbf{Combining Learning with Trajectory Optimization:}
Trajectory optimization has been combined with learning in various ways. For example, previous work has proposed to use 
imitation learning to learn the hyperparameters for a motion planner~\cite{bhardwaj2020differentiable, li2016learning, amos2018differentiable} . 
Vlastelica et al. \cite{vlastelica2021risk} combine motion planning and RL to differentiate epistemic and aleatoric uncertainty in a probabilistic setting with safety constraints. Schrum et al. \cite{schrum2022meta} combine meta-learning into trajectory optimization to increase its adaptability.
Several other works have proposed to combine RL and motion planning similar to our work~\cite{xia2021relmogen, jiang2018integrating}.
In contrast to these approaches, our method combines RL and trajectory optimization for safety-critical tasks by separating the CMDP objectives into a hierarchical structure. 

\section{Background: Constrained Markov Decision Process}
\label{sec:background}
A Constrained Markov Decision Process (CMDP)~\cite{altman1999constrained} is formulated as a tuple $(S,A,P,r,\gamma, c)$, which includes states $s \in S$, actions $a\in A$, transition function $p(s', a, s) \in P:S\times A \times S \rightarrow [0,1]$, reward functions $r$, the discount factor $\gamma$,  and the cost function $c$ that defines the safety constraint. The training objective of a CMDP is defined as 
\begin{equation}
\begin{aligned}
    &\max_{\theta} J_r(\pi_{\theta}) := \mathbb{E}_{\tau \sim \pi_{\theta}}[\sum_{t=0}^{\infty}\gamma^t r(\mathbf{s}_t, \mathbf{a}_t)] \\
    &\text{s.t. } J_c(\pi_{\theta}) := \mathbb{E}_{\tau \sim \pi_{\theta}}[\sum_{t=0}^{\infty}\gamma^t c(\mathbf{s}_t, \mathbf{a}_t)] \le C, \\
\end{aligned}
\end{equation}
where $C$ is a cost threshold, $\pi_{\theta}$ is the policy with parameters $\theta$, and $\tau$ denotes a trajectory.

\section{Problem Statement and Assumptions}
\label{sec:problem_statement}
In this work, we focus on the specific problem in which the safety constraints are defined by obstacles that we want to avoid. 
Specifically, 
all constraints are of the form: 
\begin{equation}
    ||\mathbf{s}_{t,x} - \mathbf{x}_j^{obs}|| > \epsilon \quad \forall t,j
    \label{eq:obs_constraint}
\end{equation}
where $\mathbf{s}_{t,x}$ is the location of the robot at the $t$th time step, $\mathbf{x}_j^{obs}$ is the location of the $j$th obstacle, and $\epsilon$ is a safety margin that determines how far the robot needs to stay away from the obstacles.  We also assume access to a sensor that allows us to obtain noisy measurements of the location of the obstacles $\mathbf{x}_j^{obs}$. 
Additionally, in our environments, some objects might be acceptable to interact with (such as a box that we want the robot to push) and others will occur a collision cost $c(\mathbf{s}_t, \mathbf{a}_t)$; we assume that the robot knows the type of each object and whether there will be a collision cost for interacting with that object. In addition, we assume that the obstacles are static; we leave the extension of our framework to dynamic obstacles for future work. Despite these assumptions, solving such a CMDP is still a challenging task because of the difficulty in balancing the objective with the constraints,  the difficulty of long-horizon reasoning, and the difficulty of reasoning about contacts such as safely pushing a box to a goal while avoiding obstacles. 
\section{Method: Reinforcement Learning with Safety-embedded MDP}
\label{sec:method}
\textbf{Overview:} We propose a method that combines reinforcement learning and trajectory optimization in a hierarchical structure. Instead of training an RL policy in the original action space, we propose to learn the policy in a modified action space defined by the parameters of a trajectory optimizer. Using the parameters from the policy output, the trajectory optimizer will plan a path for the agent while taking into account the safety constraints. The optimized trajectory will then be sent to a trajectory-following module that chooses robot actions to follow the path. 

Our method consists of three layers: A high-level RL agent that outputs parameters for the trajectory optimizer, a mid-level trajectory optimizer that outputs a safe trajectory, and a low-level trajectory-following module that executes the trajectory. A summary of our method is shown in Fig.~\ref{fig:method} and the pseudo-code is shown in Alg.~\ref{alg} in Appendix.~\ref{app_sec:alg}. 
We will explain below how this approach significantly reduces safety violations while also enabling our agent to learn contact-rich policies.

\begin{figure}
\centering
\includegraphics[width=0.48\textwidth]{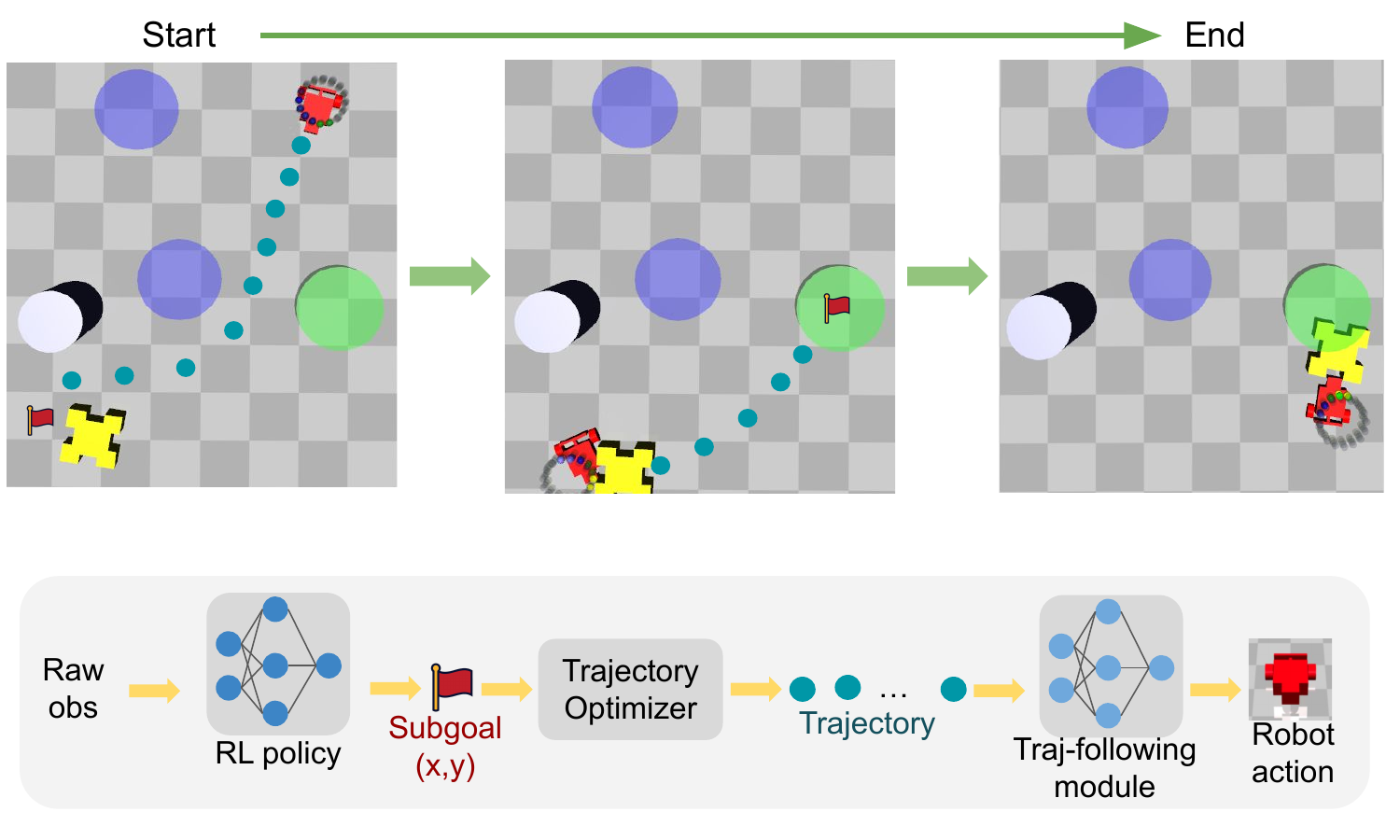}
\caption{An illustration of our method: in the Safety Gym Push task, the objective of the agent (red) is to push the box (yellow) to a goal (green) while avoiding obstacles (purple). 
Our method embeds safety constraints into the low-level trajectory optimizer to generate a safe trajectory (the dark green dots) leading toward the subgoal. The high-level RL policy outputs a subgoal (the red flag). The RL policy continually updates the subgoal output to achieve the task. }
\label{fig:method}
\end{figure}

\subsection{Safety-Embedded Markov Decision Process}
In order to optimize the CMDP (Section~\ref{sec:background}), we propose to train an RL policy over a ``Safety-Embedded Markov Decision Process" (SEMDP).
In the SEMDP, the state space $S$ 
remains the same as in the original MDP. We define a modified action space $A'$ to be a set of parameters that will be input into the trajectory optimizer. Specifically, we use a subgoal position for the agent as the action space of the RL agent in our experiments, which is the desired location of the “root node” of the agent 
(see Appendix~\ref{app_sec:root} for the definition of the ``root node").
The trajectory optimizer (described below) will then find a safe trajectory for the agent to reach the subgoal.

Based on this new RL action space $A'$ (defined as a subgoal or parameters of the trajectory optimizer), we define a new transition function $P': S\times A' \times S \rightarrow [0,1]$ which depends on the trajectory optimizer and the trajectory following module. Given the current state $\mathbf{s}_t$ and action $\mathbf{a}'_{t} \in A'$, 
the trajectory optimizer (details below) will plan a safe trajectory to reach the subgoal.  The trajectory-following module will then take $k$ actions in the original MDP to follow the trajectory.  Thus, the state $\mathbf{s}_{t+1}$ that is reached after taking action $\mathbf{a}'_t$ 
depends on the operation of the trajectory optimizer and the trajectory-following module. From the perspective of the RL agent, this transition is recorded as the tuple $(\mathbf{s}_t, \mathbf{a}'_t, \mathbf{s}_{t+1})$.  Because the SEMDP is operating over $k$ time steps in the original MDP, the reward function $r'(\mathbf{s}_t, \mathbf{a}_t)$ is modified to be the accumulated reward over $k$ steps.

Importantly, the SEMDP does not need to be a CMDP, e.g. it does not include an explicit cost constraint. This is because the cost is accounted for in the modified transition function, which uses a trajectory optimizer to find a safe trajectory to reach the subgoal. If the trajectory optimizer finds a safe trajectory and if the trajectory-following module correctly follows the trajectory, then all states visited by the agent will be safe (i.e. they will have 0 costs). As we will see, this change makes the RL optimization significantly easier. Because the SEMDP does not need an explicit cost constraint, we train the RL agent in the SEMDP with a standard method for model-free reinforcement learning, i.e. SAC~\cite{haarnoja2018soft} (details in Appendix~\ref{app_sec:details}) 

\subsection{Trajectory Optimizer}
\label{sec:method_traj_opt}
The goal of the trajectory optimizer is to find a safe and feasible trajectory to reach the subgoal $\mathbf{a}'_t$. 
In this work, we discretize the trajectory into $N$ waypoints, denoted as $\mathbf{X} := \{\mathbf{x}_1, \mathbf{x}_2, ..., \mathbf{x}_N\}$, in which $\mathbf{x}_i$ defines the position of the ``root node" of the agent. We also define the velocity at each waypoint as $\mathbf{V} := \{\mathbf{v}_1, \mathbf{v}_1, ..., \mathbf{v}_N\}$. 


Mathematically, we define the following constrained optimization problem for the trajectory optimizer:
\begin{equation}
\begin{aligned}
 \min_{\mathbf{X}, \mathbf{V}} f_{goal}(\mathbf{X}, \mathbf{a}_t') \quad
\text{s.t.} \quad 
&h_{init}(\mathbf{X},\mathbf{s}_{t,x}) \leq \delta_{init}\\
&h_{smooth}(\mathbf{X}, \mathbf{V}) \leq \delta_{smooth} \\
&\sum_{i,j}  h_{cost}(\mathbf{x}_i,\mathbf{x}_j^{obs}) \leq 0,
\label{equ:constraint_opt}
\end{aligned}
\end{equation}
in which $\textbf{s}_{t,x}$ is the position of the root node of the agent at time step $t$ and 
$\delta_{init}$ and $\delta_{smooth}$ are constants that define the constraint limits. We define each component of this optimization problem below:

\textbf{Subgoal-reaching Objective:} The optimization objective encourages the final waypoint of the trajectory to align with the subgoal location $\mathbf{a}'_{t}$ that was output by the RL policy: 
    $f_{goal}(\mathbf{X}, \mathbf{a}'_{t}) := ||\mathbf{x}_N - \mathbf{a}'_{t}||^2$
in which $||\cdot||$ denotes the $L2$ distance. Note that subgoal reaching is in the objective of this optimization but is not enforced as a constraint. Thus, occasionally the trajectory optimizer will fail to find a trajectory that reaches the subgoal $\mathbf{a}'_t$ in order to satisfy the safety constraints.

\textbf{Initial position Constraint:} The first constraint enforces that the initial waypoint needs to be located at the current position of the root node of the robot, $\mathbf{s}_{t,x}$. The corresponding cost function is defined as: 
    $h_{init}(\mathbf{X},{s}_{t,x}) := ||\mathbf{x}_1 - \mathbf{s}_{t,x}||^2$.

\textbf{Smoothness Constraint:} The second constraint enforces that the trajectory must be smooth. We assume that a sufficiently smooth trajectory can be followed by the robot; we leave for future work to incorporate a robot-specific feasibility function based on the robot dynamics.
We optimize for the location of the waypoints and the corresponding velocity at these waypoints. Non-smooth locations and changes in the velocities are penalized. The smoothness cost is defined as:

    $h_{smooth}(\mathbf{X}, \mathbf{V}) := \sum_{i=1}^{N-1} \left\| \begin{array}{c}
        \mathbf{x}_{i+1} - \mathbf{x}_i - \mathbf{v}_i\Delta t  \\
         \mathbf{v}_{i+1} - \mathbf{v}_i 
    \end{array}\right\|_K^2$
in which $||\cdot||_K$ is the Mahalanobis distance with a metric given by $K$ and $\Delta t$ is the time interval between two adjacent waypoints; this smoothness cost is derived from a constant velocity GP prior with an identity cost-weight; see prior work~\cite{barfoot2014batch,mukadam2018continuous} for details.

\textbf{Collision-avoidance Constraint:} The last set of constraints enforces that the trajectory needs to avoid collisions with obstacles. The cost of the $i$th waypoint with the $j$th obstacle is defined as:
    \begin{equation}
        h_{cost}(\mathbf{x}_i,\mathbf{x}_j^{obs}) :=\left\{ \begin{array}{cc}
             0 &\text{if } d_{i,j}>\epsilon' \\
             (\epsilon' - d_{i,j})^2& \text{otherwise} 
        \end{array}\right. , 
    \end{equation}
    in which $\mathbf{x}_j^{obs}$ denotes the location of the $j$th obstacle, $d_{i,j} := ||\mathbf{x}_i - \mathbf{x}_j^{obs}||$ denotes the distance between the $i$th waypoint and the $j$th obstacle, and $\epsilon'$ denotes a distance threshold. We choose $\epsilon'$ such that $\epsilon' \geq \epsilon$, in which $\epsilon$ is the distance threshold specified by the problem definition in Equation~\ref{eq:obs_constraint}, 
    to account for perception noise and errors in the trajectory-following module. 

We solve the constrained optimization problem in Equation~\ref{equ:constraint_opt} using the method of dual descent:
\begin{equation}
    \max_{\boldsymbol{\lambda} \geq 0} 
    \min_{\mathbf{X}, \mathbf{V}} 
    f_{goal}(\mathbf{X}, \mathbf{a}_t') + \boldsymbol{\lambda}^T 
    \left(
    \begin{array}{c}
    h_{init}(\mathbf{X},{s}_{t,x}) - \delta_{init}\\
    h_{smooth}(\mathbf{X}, \mathbf{V}) - \delta_{smooth} \\
    \sum_{i,j}  h_{cost}(\mathbf{x}_i,\mathbf{x}_j^{obs})
    \end{array}
    \right) .
\label{equ:lag}
\end{equation}
The inner loop is optimized using a trajectory optimizer; in practice, we use the Levenberg-Marquardt algorithm~\cite{wright1999numerical} implemented in Theseus~\cite{pineda2022theseus}.
The outer loop is optimized using gradient descent on $\boldsymbol{\lambda}$. Please see Appendix~\ref{app_sec::traj_opt_details} for more implementation details about the trajectory optimizer.

\subsection{Trajectory-Following Module}
The trajectory optimizer outputs a set of waypoint locations; we ignore the velocities output by the trajectory optimizer in the trajectory-following module, since their purpose was only to define the smoothness cost $h_{smooth}(\mathbf{X}, \mathbf{V})$. 
Next, we use a trajectory-following module that operates in the original robot action space to track the waypoints.
Given the trajectory $\mathbf{X} := \{\mathbf{x}_1, \mathbf{x}_2, ..., \mathbf{x}_N\}$, the trajectory-following module selects the next waypoint $\mathbf{x}_i$ and inputs the waypoint to the goal-following agent to generate low-level robot actions $a_t = \pi_{\phi}(\mathbf{s}_{t}, \mathbf{x}_i)$.

Our overall system is agnostic to the form of the goal-following agent; in our experiments, we train the goal-following agent using reinforcement learning in an obstacle-free environment with only the robot and a randomly sampled goal. The goal-following agent is goal-conditioned $\pi_\phi(\mathbf{s}_{t}, \mathbf{g})$ and is trained to reach a goal $\mathbf{g}$ that is randomly sampled around the robot.
More implementation details about the trajectory-following module are in Appendix~\ref{app_sec:traj_follow}.

\section{Experiments}
\label{sec:experiment}
We evaluate our method on Safety Gym simulation benchmarks~\cite{ray2019benchmarking}. We also transfer the policy to a real-world task of pushing a box around obstacles to a goal in Sec.~\ref{sec:real-robot}. 
\subsection{Safety Gym Setup}
Safety Gym~\cite{ray2019benchmarking} is a set of benchmark environments that can be used to evaluate methods under a CMDP framework.
In our experiments, we focus on the challenging 
``Push" tasks of Safety Gym, in which the robot has to push a box towards a goal and avoid obstacles. The Push tasks require reasoning about rich contacts between the robot and the environment, while also reasoning about safety; this environment is challenging for previous methods, which would run into obstacles (high cost) or get stuck and cannot finish the task (low reward). Previous work on Safety Gym used a cost threshold of 
C=25~\cite{ray2019benchmarking}; in contrast, we use a stricter cost threshold of C=0 in our experiments for purposes of evaluation. We evaluate our method with four different robot morphologies: Point, Car, Mass, and Ant. Please refer to Appendix~\ref{app_sec:safety_gym} for details.



We compare our method to the state-of-the-art safe RL methods: CPO~\cite{achiam2017constrained}, PPO-Lagrangian (PPO Lag), TRPO-Lagrangian~\cite{ray2019benchmarking} (TRPO Lag), Safety Editor~\cite{yu2022towards} (SE) and Constrained Variational Policy Optimization~\cite{liu2022constrained} (CVPO) using the author-provided code. Additionally, we also compare to a safe exploration method~\cite{dalal2018safe}, whose results are shown in Appendix.~\ref{app_sec:safe_expl_exp}.
Four seeds are used for each method during training.

\subsection{Safety Gym Results}
\begin{figure*}
    \centering
    \includegraphics[width=0.92\textwidth]{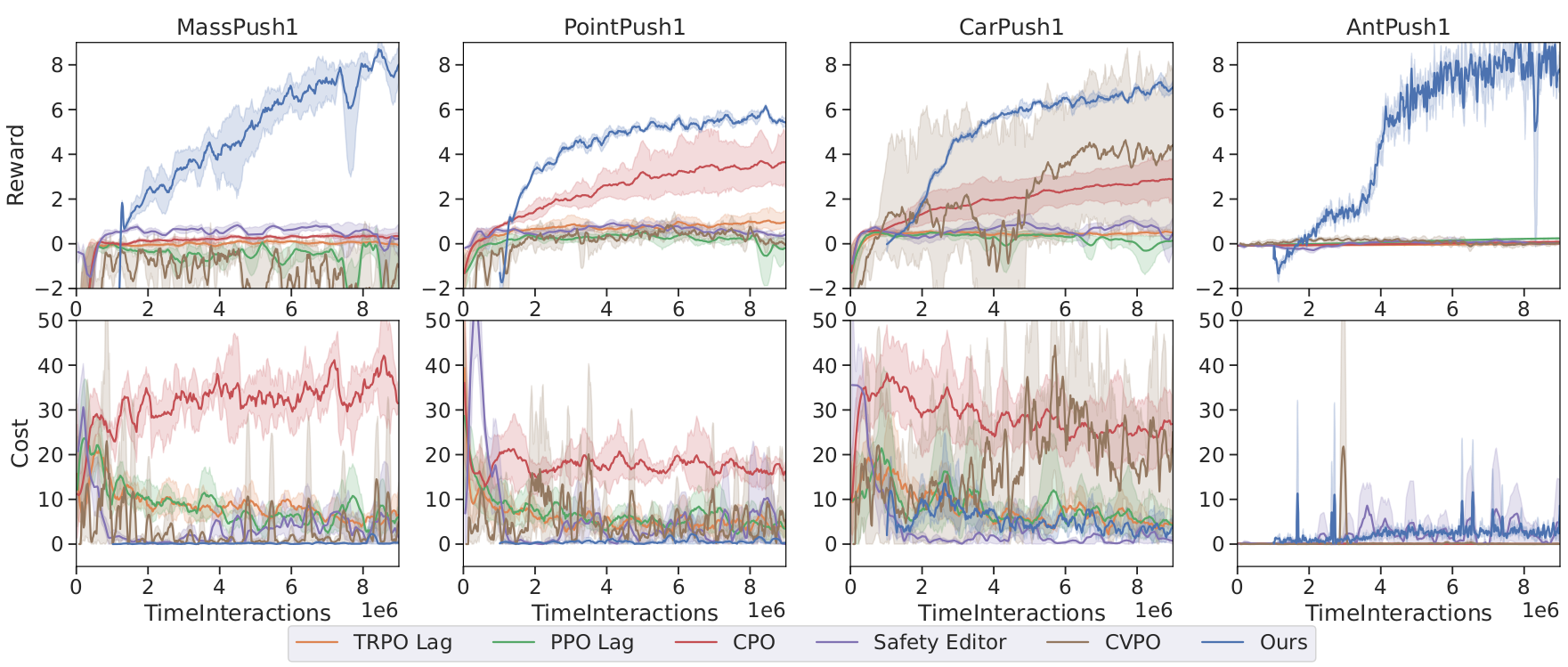}
    \caption{Training curves of our method compared to the baseline methods. The shadow region denotes the standard error of different seeds. Our method starts from $1\mathrm{e}{6}$ steps instead of $0$ to denote the training of the goal-reaching policy. In these experiments, the cost is defined as the total number of time steps for which the agent violates the safety constraints within an episode.
    Our method achieves a lower cost than the baselines. It still incurs some cost during training because, during training time, we are using a fixed Lagrangian parameter for computation reasons and to encourage exploration.  }
    \label{fig:baseline}
\end{figure*}

\label{sec:safety_gym_exp}
The results during training are shown in Fig.~\ref{fig:baseline}. The results in more difficult level 2 tasks PointPush2 and CarPush2 are shown in Fig.~\ref{fig:addtional_training_curves} in Appendix.~\ref{app_sec:add_training_curves}.
We smooth the curves for better visualization.
As shown, our method achieves a much higher reward than the baselines with very little incurred cost. 
We use a fixed $\boldsymbol{\lambda}$ during training to speed up computation and to encourage exploration, which leads to some safety violations during training; at test time, we adjust $\boldsymbol{\lambda}$ to optimize Equation~\ref{equ:lag}, leading to fewer safety violations.



We evaluate the converged policy at the final iteration of training. Mean actions are chosen instead of sampled actions from the policy. Each policy is evaluated for 50 episodes and the average results are shown in Table~\ref{tab:eval}. An additional analysis showing reward rather than success rate as the metric is shown in Table~\ref{app_tab:eval} in Appendix.~\ref{app_sec:eval_with_reward}. The number of safety errors from our method is reduced compared to training because we adjust $\boldsymbol{\lambda}$, unlike in training when $\boldsymbol{\lambda}$ is held fixed.


As noted previously, in prior work on Safety Gym, a cost threshold of 25 was used~\cite{ray2019benchmarking}; in our experiments, we use a stricter cost threshold of 0.  This leads to significantly worse performance for the Lagrangian methods, which are unable to achieve a reasonable reward due to training instability. 

\begin{table*} \small
\begin{center}
\caption{Evaluation results of the final converged policies; 
see text and Appendix.~\ref{app_sec:training_details} for details. Experiments with a cost exceeding 10 are marked in gray to indicate that they are not safe. See Table~\ref{app_tab:eval} for the reward instead of the success rate.
}
\begin{tabular}{|c|c|c|c|c|c|c|c|} 
 \hline
  & & SEMDP (ours)  & CPO~\cite{achiam2017constrained} & PPO Lag~\cite{ray2019benchmarking} & TRPO Lag~\cite{ray2019benchmarking} & SE\cite{yu2022towards} & CVPO\cite{liu2022constrained}\\ 
   \hline
MassPush1 & success rate & \textbf{0.55}  & \textcolor{gray}{0.11} & 0.01 & 0.05& 0.02 & 0\\ 
 \cline{2-8}
 & cost & \textbf{0.00} & \textcolor{gray}{28.00} & 1.41 & \textbf{0.00} & 3.01 & 0.80\\ 
 \hline

 PointPush1 & success rate & \textbf{0.84} & 0.77 & 0.00 & 0.08 & 0.03 & 0.00\\ 
 \cline{2-8}
 & cost & \textbf{0.00} &5.04 & 8.34 & 1.39 & 4.03 & 4.90 \\ 
 \hline

 CarPush1 & success rate & \textbf{0.88} & \textcolor{gray}{0.83} & 0.02 & 0.11 & 0.05 & \textcolor{gray}{0.01} \\ 
 \cline{2-8}
 & cost & \textbf{0.00} & \textcolor{gray}{14.44} & 2.28 & 3.64 & 0.47& \textcolor{gray}{23.5}\\ 
\hline
 
 AntPush1 & success rate & \textbf{0.79} & 0.02& 0.00 & 0.00 & 0.00 & 0.00\\ 
 \cline{2-8}
 & cost & 0.48 & 9.35 & \textbf{0.00} & \textbf{0.00} & \textbf{0.00} & \textbf{0.00} \\ 
\hline

PointPush2 & success rate & \textbf{0.57} & \textcolor{gray}{0.40} & 0.03 & 0.02 & 0.01 & \textcolor{gray}{0.00} \\ 
 \cline{2-8}
 & cost & \textbf{0.00} & \textcolor{gray}{27.40} & 4.81 & 4.73 & 0.44 & \textcolor{gray}{17.60}\\ 
\hline

CarPush2 & success rate & \textbf{0.58} & \textcolor{gray}{0.38} & \textcolor{gray}{0.01} & 0.00 & 0.01 & \textcolor{gray}{0.00} \\ 
 \cline{2-8}
 & cost & \textbf{0.25} & \textcolor{gray}{41.58} & \textcolor{gray}{54.02} & 7.28 & 1.29 & \textcolor{gray}{43.62} \\ 
\hline
\end{tabular}
 \label{tab:eval}
\end{center}
\end{table*}


\subsection{Ablations and Additional Analysis}
\label{sec:ablation}

\begin{table}
    
\small
\begin{center}
\caption{Evaluation results of our method and ablations. 
Each method was trained for $1e7$ environment interaction steps.  Experiments with a cost exceeding 10 are marked in gray to indicate that they are not safe. 
}
\begin{tabular}{|c|c|c|c|} 
 \hline
  & & SEMDP (ours) &  SAC + PPO Lag \\ 
   \hline
MassPush1 & reward & \textbf{4.31} &\textcolor{gray}{14.62} \\ 
 \cline{2-4}
 & cost & \textbf{0.00} &\textcolor{gray}{40.25} \\ 
 \hline

 PointPush1 & reward & \textbf{5.69}  & \textcolor{gray}{-0.87}\\ 
 \cline{2-4}
 & cost & \textbf{0.00} & \textcolor{gray}{24.16}\\ 
 \hline

 CarPush1 & reward & \textbf{4.57} & 0.18\\ 
 \cline{2-4}
 & cost & \textbf{0.00}& 6.75\\ 
 \hline
\end{tabular}
 \label{tab:ablation}
\end{center}
\end{table} 

We perform additional ablation experiments to understand the reason behind our method's strong performance.

\textbf{How much of our improvement over the baselines is attributed to using a learned trajectory-following module?}  First, note that 
we do not train a trajectory-following module for the Mass agent, since we can directly command this agent to any local delta position using its low-level action space. 
As shown in the ``MassPush1" experiments in Fig.~\ref{fig:baseline} and Table~\ref{tab:eval}, our method still significantly outperforms the baselines.  This demonstrates that the benefits of our method are not from using a learned trajectory-following module.  We believe that the benefits come from training an RL agent in a Safety-Embedded MDP defined by a safe trajectory optimizer. We also perform an experiment in which we modify PPO Lagrangian to incorporate a trained goal-reaching low-level agent (which we still outperform); see details in Appendix~\ref{app_sec:ablation_traj_following}.


\textbf{Do we need a trajectory optimizer?}
In this ablation, we attempt to replace the safe trajectory optimizer with a learned ``safe" goal-reaching policy. 
Instead of using an optimization-based trajectory optimizer, we use PPO Lagrangian to train a low-level ``safe goal reaching" policy 
with a reward of reaching a randomly sampled goal and a cost constraint of avoiding obstacles. The high-level policy is trained with SAC, the same as in our method, to output subgoals for the low-level goal-reaching agent. 
The intention of this experiment is to be as similar to our method as possible but replace the trajectory optimizer with a low-level goal-reaching agent trained with safe RL. 
The results of this experiment can be found in Table~\ref{tab:ablation}, referred to as ``SAC + PPO Lag." 
As can be seen, this method also performs poorly, demonstrating that a safe trajectory optimizer is needed to ensure safety; training a cost-aware low-level agent with PPO Lagrangian is not sufficient to obtain safe performance.

\subsection{Real-Robot Experiments}

\begin{figure*}[htbp]
    \centering
    \includegraphics[width=\textwidth]{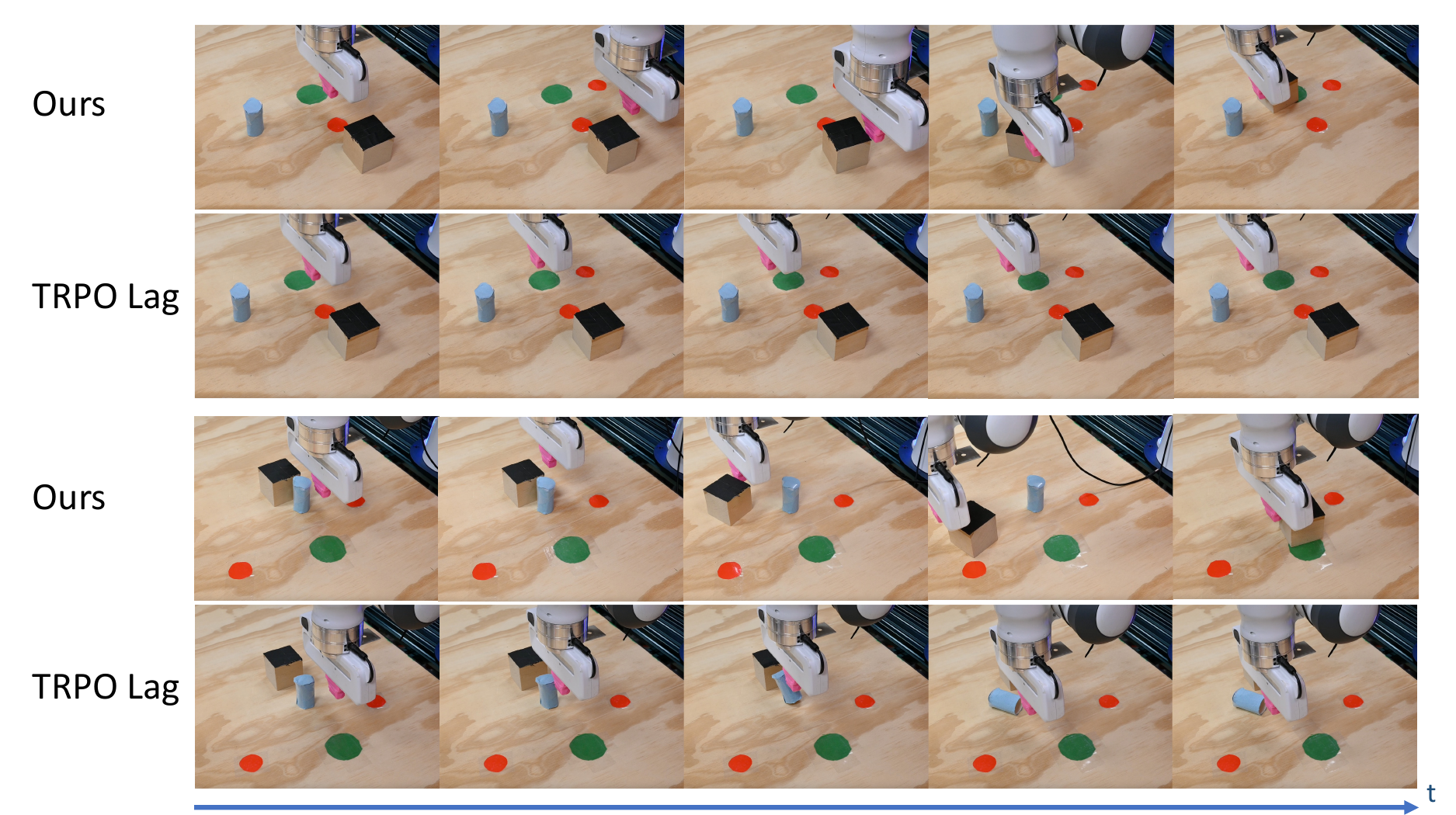}
    \caption{We set up a real-robot environment similar to the Safety Gym Push task. The fingertip of the Franka robot (pink) is used to push the box (black) toward the goal (green). It needs to avoid hazards (red) and avoid getting stuck at the pillar (blue). Each row shows four frames of a single episode. 
    We compare our method with TRPO Lagrangian, which has the best performance among the baselines based on the simulation experiments. }
    \label{fig:robot_setup}
\end{figure*}
\label{sec:real-robot}
We use a real-robot version of the Push task for evaluation, using a Franka Panda gripper. The details of the experiments are shown in Appendix.~\ref{app_sec:real_robot}. In this experiment, the fingertip of the gripper moves in a plane to push the box toward the goal. As in the simulation setup, the robot also needs to avoid hazards and try not to get stuck by the pillar.
The real robot experiment is shown in Fig.~\ref{fig:robot_setup}. An episode is considered successful if the robot is able to push the box into the goal region within 60 time steps.

\begin{table}
\small
    \caption{Results of the real robot experiment. While both methods have 0 cost in the real world, our method achieves a higher success rate and reward than the baseline.}
    \centering
    \begin{tabular}{|c|c|c|c|}
    \hline
    Method & Succ rate & reward & cost \\
        \hline
        SEMDP (ours) & \textbf{8/10} & \textbf{1.33}& \textbf{0}\\
        \hline
        TRPO Lag & 0/10 & 0.22& \textbf{0} \\
        \hline
    \end{tabular}

    \label{tab:real_robot}
\end{table}

We compare our method with TRPO Lagrangian. We evaluate each method with 10 different layouts; evaluating each layout with both our method and the baseline.  The results are shown in Table~\ref{tab:real_robot}. Both methods are safe in the real world, but our method has a much higher success rate and reward. The reasons for the failure of our method include timing out or getting stuck around the obstacles. For the TRPO Lagrangian baseline, the robot is able to move towards the box but is not successful in pushing the box to the goal, which also matches its performance in simulation.
\section{Limitations and Conclusions}
The main limitations of our system are the assumptions mentioned in Sec.~\ref{sec:problem_statement}. Further, even with a safe trajectory optimizer,  
it is still hard to guarantee safety in practice, due to perceptual errors or modeling inaccuracies. 
In our case, we also use a learned trajectory-following module which might not follow the trajectory perfectly; a model-based trajectory optimizer that takes into account the agent dynamics could be used here to ensure feasible trajectories. We leave such an extension of our method for future work.

In conclusion, we propose a hierarchical framework, in which the RL agent optimizes the reward in a modified MDP which is embedded with a trajectory optimization algorithm to ensure safety. 
We test our method on Safety Gym benchmarks and a real-robot pushing task, demonstrating better performance than the baselines in terms of both rewards and costs. 
In future work, our framework can be generalized in that  
the RL agent can output any parameters that define the objective for the trajectory optimizer, and the trajectory optimizer can take any form as long as it is compatible with the output of the RL policy. We believe our work will contribute to the field of safe robot learning by demonstrating the importance of combining RL and trajectory optimization in safety-constrained optimization tasks.  

\bibliographystyle{IEEEtran}
\bibliography{reference}
\clearpage

\newpage
\appendix
\addcontentsline{toc}{section}{Appendix} 


\section{Additional Experiments}
\label{app_sec:additional_exp}
\subsection{Evaluation with Reward}
\label{app_sec:eval_with_reward}
In Sec.~\ref{sec:safety_gym_exp}, we evaluated each method in terms of success rate. In Table~\ref{app_tab:eval} we present an additional analysis  in terms of the sum of rewards in each episode. Note that, to compute the success rate for Table~\ref{tab:eval}, we terminate each episode when the box reaches the goal; on the other hand, in this experiment, we continuously sample a new goal until it reaches 1000 steps. As a result, the cost in Table~\ref{app_tab:eval} is slightly different than the cost in Table~\ref{tab:eval} since the episode termination criteria is defined differently.
In the Ant agent environment, we also terminate the episode when the agent flips over and cannot recover. 

\begin{table*}[ht] \small
\begin{center}
\caption{Evaluation results of the final policy, using an adaptive Lagrangian parameter for our method. See text and Appendix.~\ref{app_sec:training_details} for details.
}

\begin{tabular}{|c|c|c|c|c|c|c|c|} 
 \hline
  & & SEMDP (ours)  & CPO~\cite{achiam2017constrained} & PPO Lag~\cite{ray2019benchmarking} & TRPO Lag~\cite{ray2019benchmarking} & SE~\cite{yu2022towards} & CVPO~\cite{liu2022constrained}\\ 
   \hline
MassPush1 & reward & \textbf{4.18}  & \textcolor{gray}{0.78} & -2.26 & -0.39& 0.36 & -1.95\\ 
 \cline{2-8}
 & cost & \textbf{0.00} & \textcolor{gray}{31.48} & 2.6 & 0.05 & 10.43 & 2.75\\ 
 \hline

 PointPush1 & reward & \textbf{5.29} & \textcolor{gray}{5.47} & -5.61 & 0.68 & 0.62 & -0.12\\ 
 \cline{2-8}
 & cost & \textbf{0.00} & \textcolor{gray}16.45 & 6.61 & 0.45 & 5.16 & \textbf{0.00} \\ 
 \hline

 CarPush1 & reward & \textbf{6.47} & \textcolor{gray}{4.4} & -2.34 & 0.45 & 0.58 & \textcolor{gray}{5.34} \\ 
 \cline{2-8}
 & cost & \textbf{0.03} & \textcolor{gray}{30.35} & 3.17 & 1.46 & 0.92& \textcolor{gray}{22.00}\\ 
\hline
 
 AntPush1 & reward & \textbf{6.55} & 0.12& 0.24 & -0.02 & 0.01 & 0.05\\ 
 \cline{2-8}
 & cost & 0.48 & 0.01 & \textbf{0.00} & \textbf{0.00} & 0.85 & \textbf{0.00} \\ 
\hline

PointPush2 & reward & \textbf{3.47} & \textcolor{gray}{2.05} & \textcolor{gray}{-7.62} & 0.18 & 0.22 & \textcolor{gray}{0.85} \\ 
 \cline{2-8}
 & cost & 0.08 & \textcolor{gray}{32.63} & \textcolor{gray}{11.95} & 4.20 & \textbf{0.01} & \textcolor{gray}{11.25}\\ 
\hline

CarPush2 & reward & \textbf{4.22} & \textcolor{gray}{2.92} & 0.24 & 0.18 & -0.55& \textcolor{gray}{-7.45} \\ 
 \cline{2-8}
 & cost & \textbf{0.29} & \textcolor{gray}{52.40} & 4.16 & 3.29 & 1.49 & \textcolor{gray}{51.13} \\ 
\hline
\end{tabular}
 \label{app_tab:eval}
\end{center}
\end{table*}

\subsection{Time Used for Inference}

Adding an additional layer of trajectory optimization does require more time during inference. For example, The average time for inference in our method is 0.083 seconds.
The inference time used in CPO, PPO Lagrangian, and TRPO Lagrangian is all 0.00017 seconds.
Specifically, in our method, the trajectory optimization module contributes most to the inference time. the trajectory optimization replans every 10 time steps, while the average time used for trajectory optimization itself is 0.76 seconds. 

The inference for the baselines mentioned above is very fast since there is no planning during the inference. Instead, observations are passed into a three-layer neural network to obtain low-dimensional actions.  

\subsection{Additional Training Curves}
\label{app_sec:add_training_curves}
\begin{figure*}
    \centering
    \makebox[\textwidth][c]{\includegraphics[width=\textwidth]{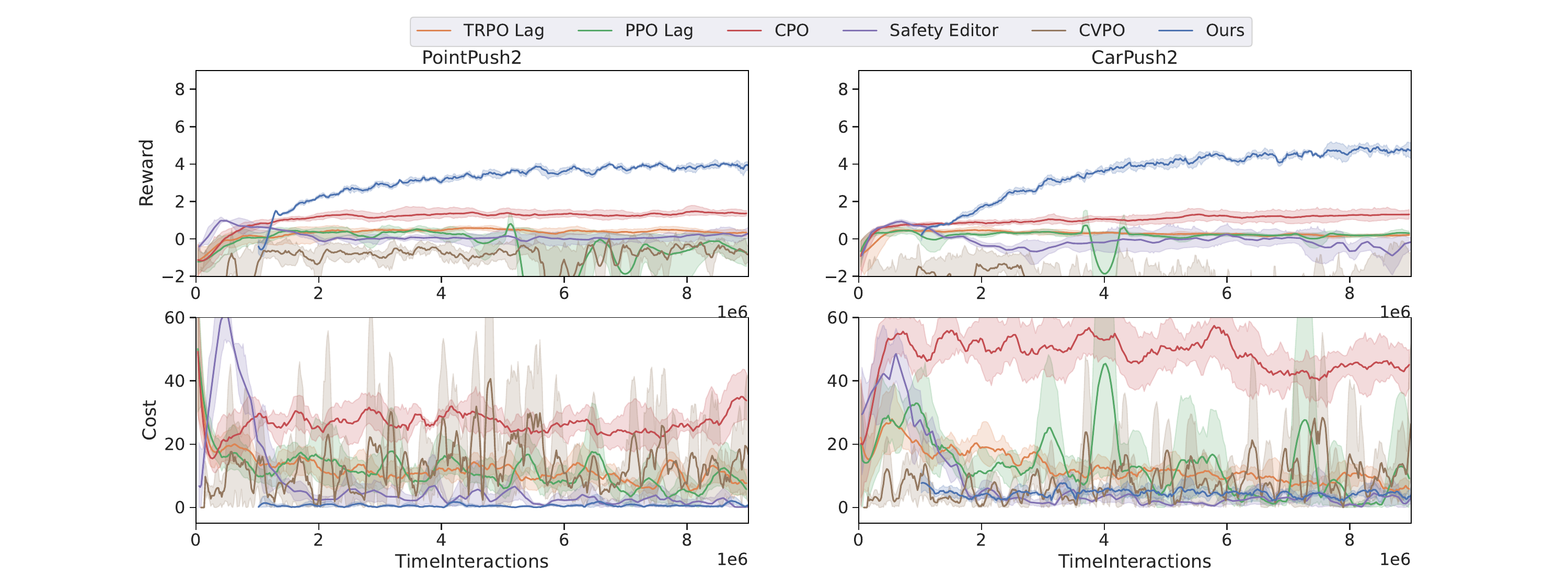}}
    \caption{Additional training curves of our method compared to the baseline methods. The shadow region denotes the standard error of different seeds. Our method starts from $1\mathrm{e}{6}$ steps instead of $0$ to denote the training of the goal-reaching policy. 
    Our method achieves a lower cost than the baselines. It still incurs some cost during training because, during training time, we are using a fixed Lagrangian parameter for computation reasons and to encourage exploration.}
    \label{fig:addtional_training_curves}
\end{figure*}

The training curves of our method and the baseline methods in PointPush2 and CarPush2 are shown in Fig.~\ref{fig:addtional_training_curves}. These two tasks are harder than PointPush1 and CarPush1 since they have more obstacles in their environment.
\subsection{Safe Exploration Experiments}
\label{app_sec:safe_expl_exp}
\begin{figure*}
    \centering
    \includegraphics[width=0.7\textwidth]{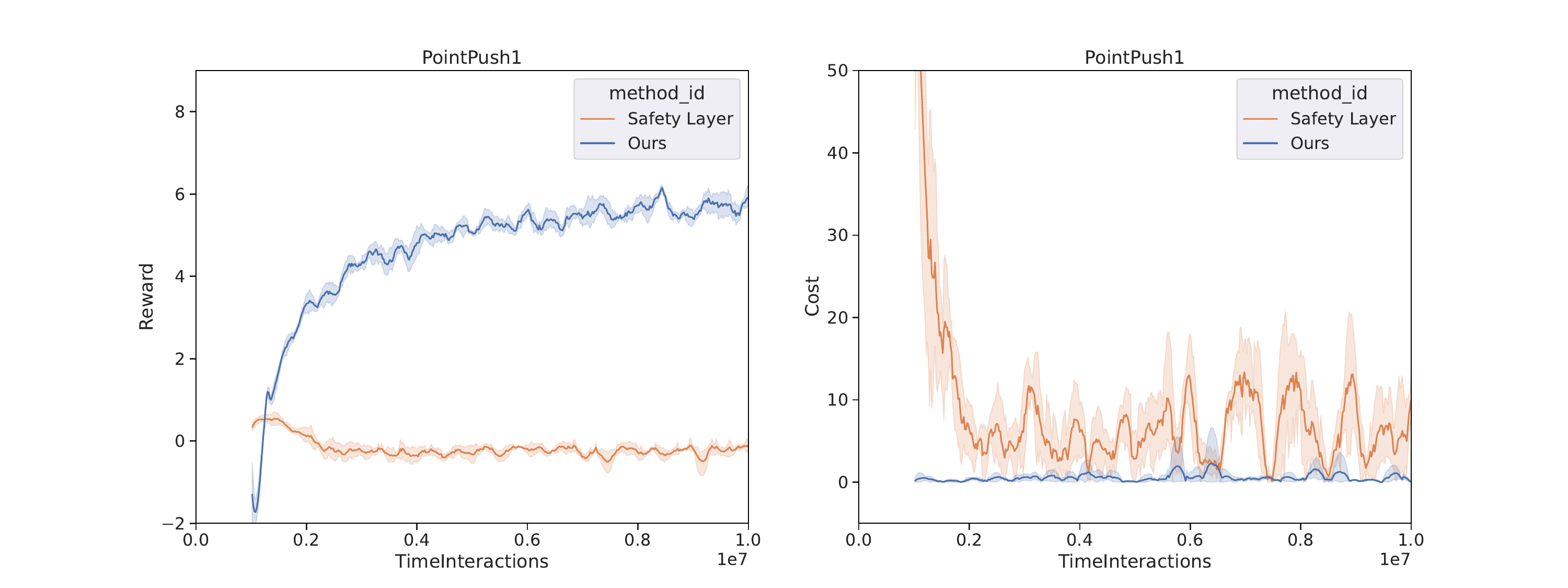}
    \caption{Comparison between our method and the Safety Layer method.~\cite{dalal2018safe}.}
    \label{fig:rebuttal_fig}
\end{figure*}

Safe exploration, which is considered another line of work in safe RL, also focuses on satisfying safety constraints for reinforcement learning agents. Compared to the safe RL methods mentioned in the main text, additional information such as an explicit dynamics models used to predict the future states given the action sequence,
is usually assumed to be given. 
Still,
we compare 
our method with the Safety Layer method~\cite{dalal2018safe}. In the Safety Layer method, we first randomly sampled actions for $1\mathrm{e}{6}$ time steps to train the cost function which is used to set up the safety layer. Then the safety layer is combined with SAC to train a safe RL agent. We compare it with our method in the PointPush1 environment. The results are shown in Fig.~\ref{fig:rebuttal_fig}, with our method shown in blue and the baseline shown in orange. This result shows that  Safety Layer  cannot outperform our method in terms of either reward or cost. It is generally hard to train an accurate cost function in practice, leading to poor performance by this method.
\subsection{Adjusting the Trade-off between Reward and Cost}
\label{sec:trade_off}
\begin{figure*}[htbp]
    \centering
    \includegraphics[width=0.8\textwidth]{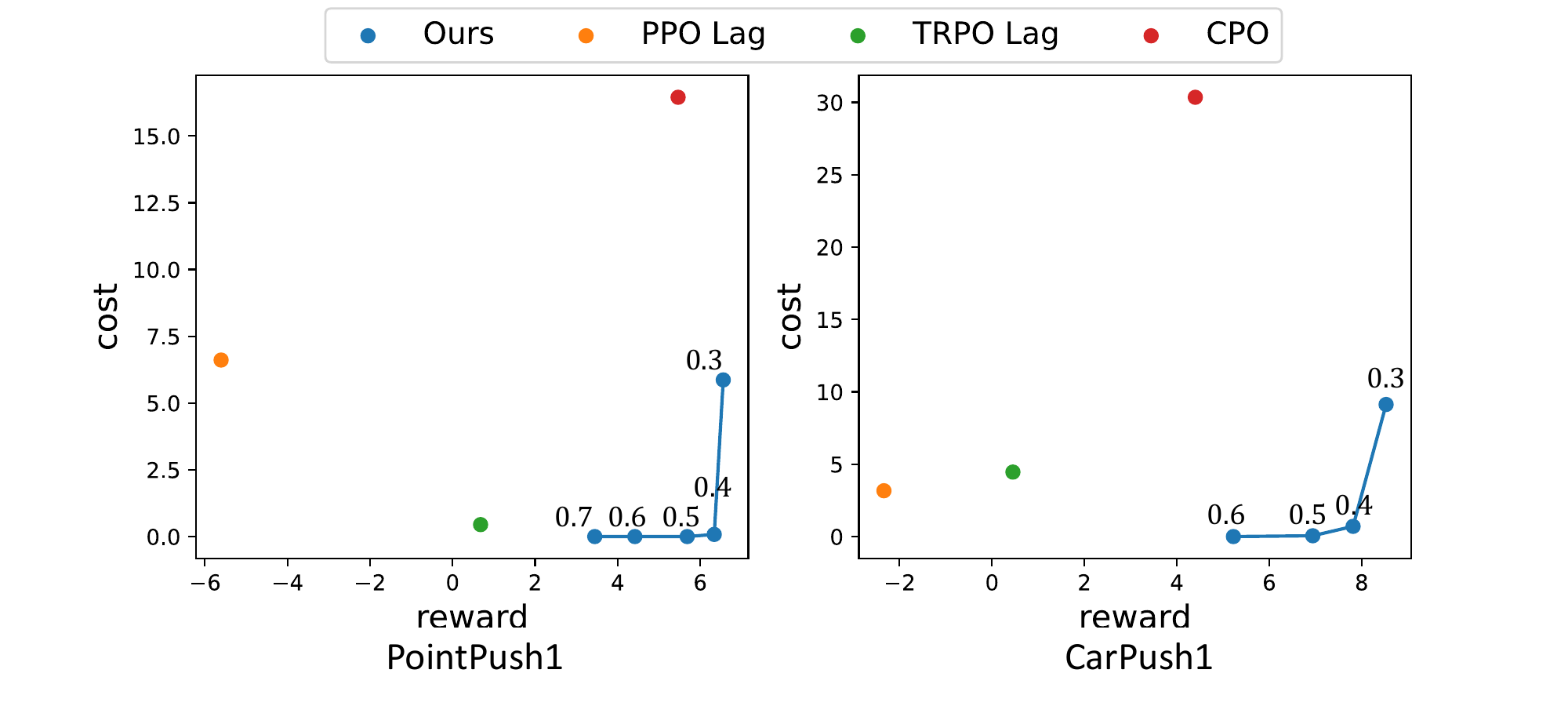}
    \caption{Trade-off between the reward and the cost for different methods. The blue dots represent our method with different $\epsilon'$. The numbers above the blue dots denote the value of $\epsilon'$.}
    \label{fig:trade_off}
\end{figure*}

We can adjust the trade-off between the reward and the cost during test time in the proposed method. Specifically, we can adjust the threshold $\epsilon'$ defined in Equation~\ref{eq:obs_constraint}. The results are summarized in Fig.~\ref{fig:trade_off}. As a comparison, we also include three baselines: CPO, PPO-Lagrangian, and TRPO-Lagrangian.

From Fig.~\ref{fig:trade_off}, smaller $\epsilon'$ might lead to a less conservative agent with higher reward and higher cost. Nonetheless, our method has the higher reward and lower cost compared to the baselines across different values of $\epsilon'$. 

\subsection{How much of our improvement over the baselines is attributed to using a learned trajectory-following module}
\label{app_sec:ablation_traj_following}
As an additional experiment to understand the effects of a trajectory-following module, we modify two of the baselines to incorporate a trained goal-reaching low-level agent. It is trained in the same way as the goal-reaching agent for our method except that the subgoal is randomly sampled to match the inference distribution. Note that this goal-reaching agent does not incorporate any safey constraints. 
Thus for this baseline, we use PPO Lagrangian to train a high-level policy that outputs a (hopefully safe) subgoal.  The results of this experiment can be found in Table~\ref{app_tab:ablation}, referred to as ``PPO Lag + SAC." The results for the Mass agent are left blank since the Mass agent does not require learning a low-level goal-reaching policy. As can be seen, ``PPO Lag + SAC" 
performs poorly, incurring a low reward and many safety violations. Though ``PPO Lag + SAC" performs slightly better than the PPO Lagrangian method, there is still a huge performance gap compared to our method. This demonstrates that the benefits of our method do not come directly from the trajectory-following module; incorporating such a module into the baselines still leads to poor performance.

\begin{table*} \small
\begin{center}
\caption{Evaluation results of our method and ablations. 
Each method was trained for $1e7$ environment interaction steps.. 
}
\begin{tabular}{|c|c|c|c|c|c|} 
 \hline
  & & SEMDP (ours) & PPO Lag + SAC  & SEMDP w/ gt\\ 
   \hline
MassPush1 & reward & \textbf{4.31}  & \diagbox[innerwidth=7em , height=\line]{}{} &10.00\\ 
 \cline{2-5}
 & cost & \textbf{0.00} & \diagbox[innerwidth=7em , height=\line]{}{} &0.00\\ 
 \hline

 PointPush1 & reward & \textbf{5.69}  & 0.16& 5.70\\ 
 \cline{2-5}
 & cost & \textbf{0.00} &1.41 & 0.00\\ 
 \hline

 CarPush1 & reward & \textbf{4.57} & 0.15& 5.33\\ 
 \cline{2-5}
 & cost & \textbf{0.00}& 5.58& 0.00\\ 
 \hline
\end{tabular}
 \label{app_tab:ablation}
\end{center}
\end{table*}
\subsection{How much is our performance affected by perceptual errors?}  
As noted, some of the errors in our system come from perceptual errors, in which the location of the obstacles is perturbed by some noise in the Safety Gym. To measure the effect of these errors, we perform an experiment in which we allow our trajectory optimizer to have access to the ground-truth location of the obstacles, while the RL agent still takes in the noisy LiDAR observations as input.  The results can be found in Table~\ref{app_tab:ablation} and it is denoted as ``SEMDP w/ gt".  As expected, the performances for all the environments have increased. The MassPush1 task has the greatest improvement in reward by switching to ground-truth locations. This is because perception errors are the major source oferrors for the Mass agent, for which the agent dynamics are relatively simple.  

\subsection{Experiments with Safety Gym Goal Tasks}
\begin{figure*}[ht]
    \centering
    \includegraphics[width=1\textwidth]{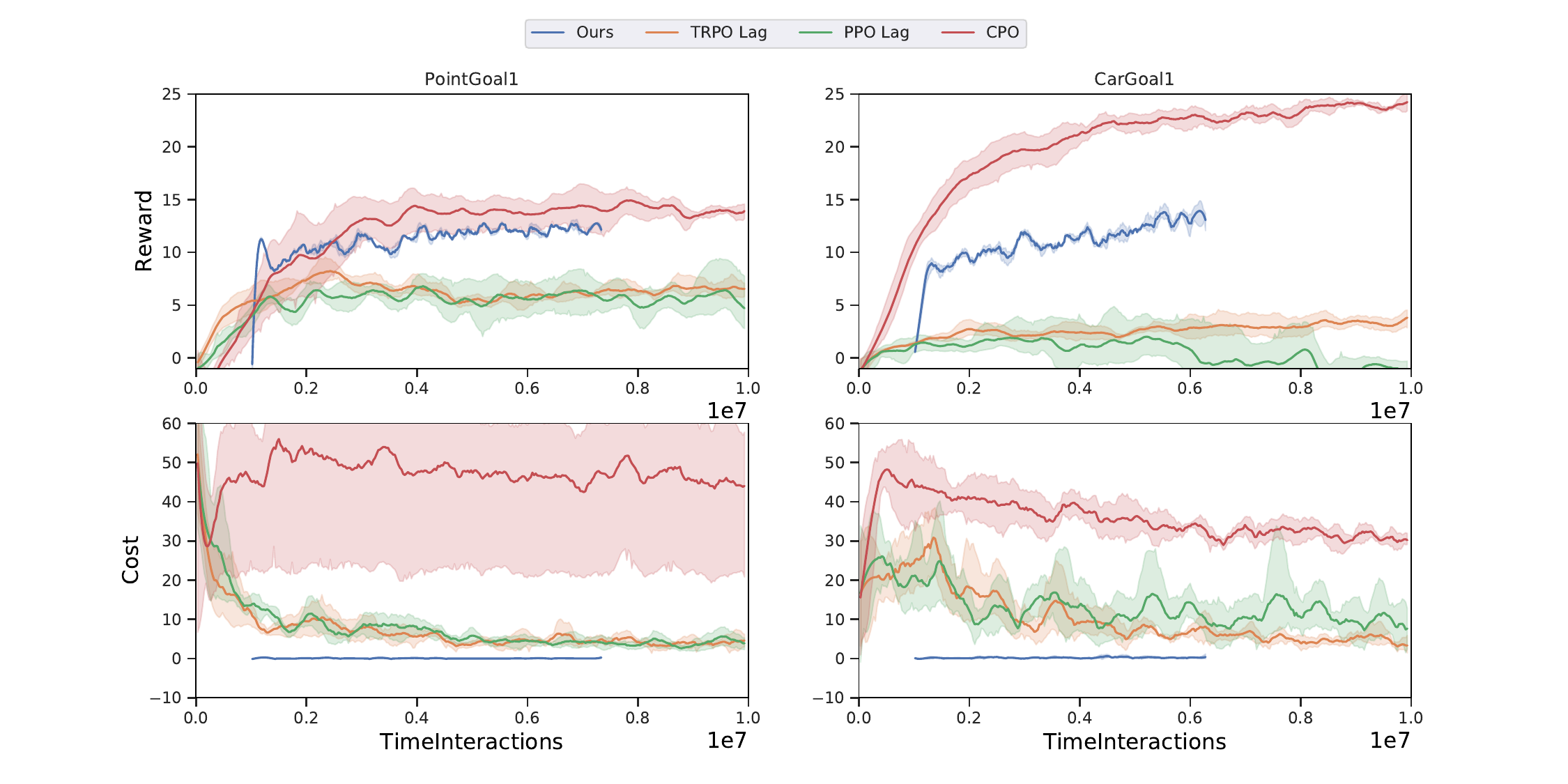}
    \caption{Training curves of our methods compared to the baseline methods in the Goal environment. The curves have been smoothed for better visualization. Our method starts from $1\mathrm{e}{6}$ steps instead of 0 to denote the training of the goal-reaching policy. Our method achieves the lowest cost among all the baseline methods. }
    \label{fig:baseline_goal}
\end{figure*}

We evaluate our method with an additional task, the ``Goal" task in SafetyGym~\cite{ray2019benchmarking}. In the Goal task, the robot itself needs to go to a specific goal and avoid obstacles instead of moving an object to the goal. The results are shown in Fig.~\ref{fig:baseline_goal}. Our method still achieves the lowest cost and a relatively high reward during training.

\subsection{Analyzing the Ablations}



In Section~\ref{sec:ablation}, we discussed an ablation called ``SAC + PPO Lag" in which we train a low-level policy to safely reach goals with PPO Lagrangian, and then we use SAC to output subgoals (see Section~\ref{sec:ablation} for further discussion).  In this section, we analyze why this method appears to perform so poorly in Table~\ref{tab:eval}.  

\begin{figure*}[ht]
    \centering
    \includegraphics[width=0.7\textwidth]{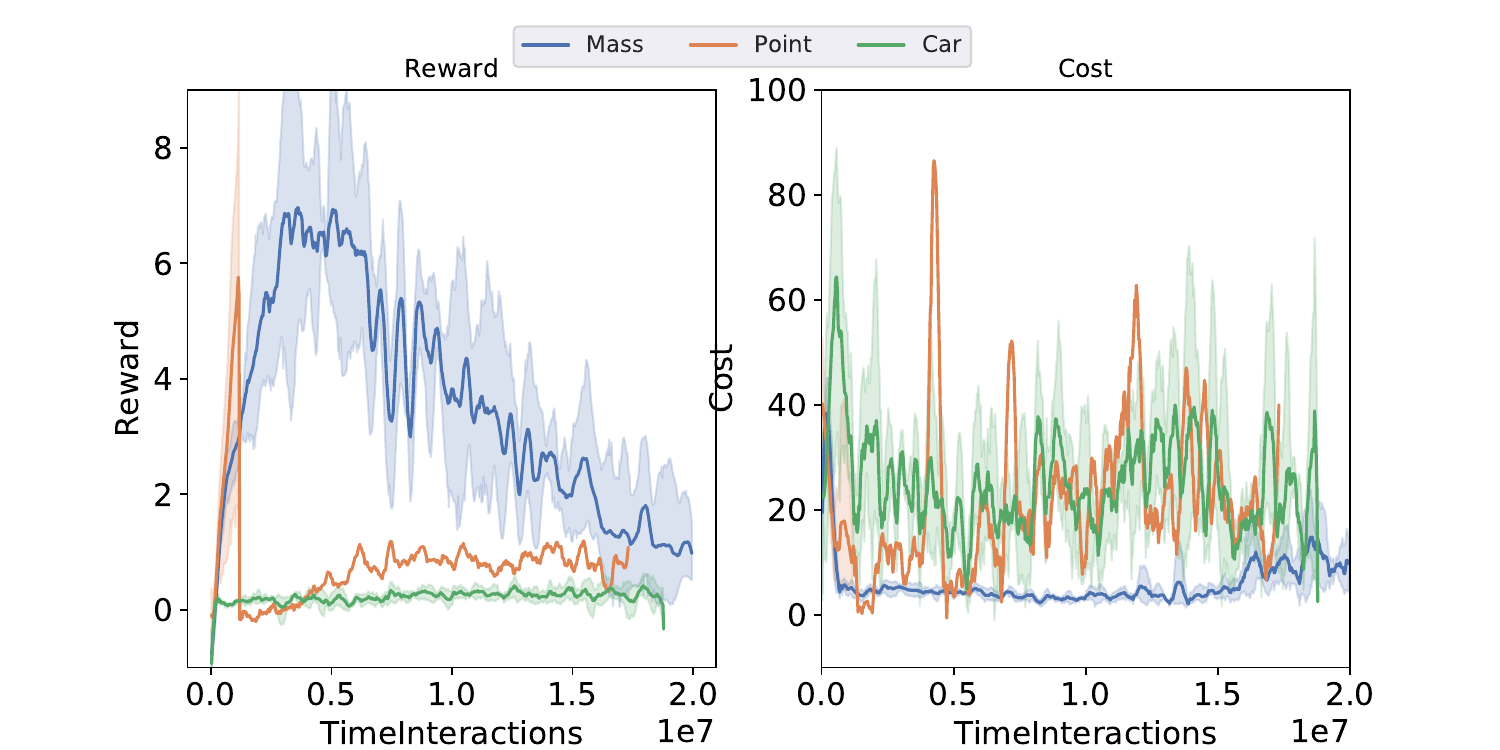}
    \caption{Training curves of the goal-reaching policies used in ``SAC + PPO Lag".}
    \label{fig:safe_low_level}
\end{figure*}

The training curves for the low-level safe goal-reaching policy trained with PPO Lagrangian are shown in Fig.~\ref{fig:safe_low_level}. As can be seen, these policies do not learn to be safe and the cost during training is always above 0. Hence, integrating such a low-level policy with a high-level SAC agent leads to poor overall performance for ``SAC + PPO Lag" baseline. This experiment highlights the difficulties of training even a safe short-horizon policy with PPO Lagrangian.

\subsection{Training Curves of the Ablations}

\begin{figure*}[ht]
    \centering
    \includegraphics[width=\textwidth]{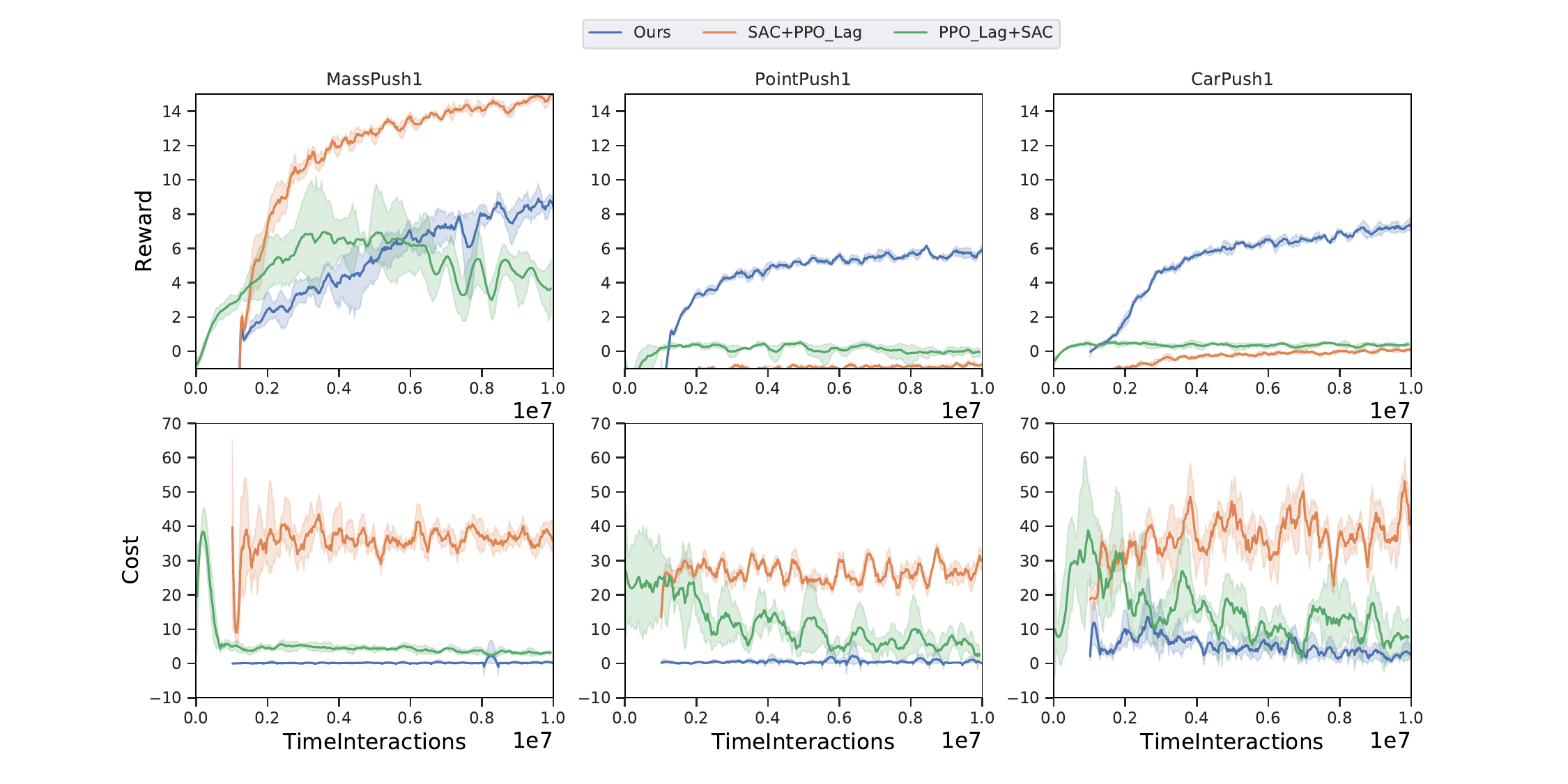}
    \caption{Training curves of our method and 2 ablations (see Section~\ref{sec:ablation} for details on these methods). }
    \label{fig:ablation}
\end{figure*}

The training curve of ablation methods are shown in Fig.~\ref{fig:ablation} which corresponds to the results in Table~\ref{tab:eval}. The ablations have much higher cost than our method during training and significantly higher cost than our method during inference (in which our method obtains 0 cost) as shown in Table~\ref{tab:eval}.

\section{Implementation Details}
\label{app_sec:details}

\subsection{Pseudocode}
\label{app_sec:alg}
The pseudocode for our method can be found in Algorithm~\ref{alg}.

\begin{algorithm}
\caption{Reinforcement Learning in a Safety-Embedded MDP with Trajectory Optimization}
\label{alg}
\begin{algorithmic}
\Require RL policy output interval $k$, goal-following agent $\pi_{\phi}$.
\State Initialize the replay buffer $D$, high-level RL policy $\pi_{\theta}$.
\For {each episode} 
\For {each environment step}
\State Select subgoal: $\mathbf{a}_t' \sim \pi_{\theta}(\mathbf{s}_t)$.
\State Input the subgoal $\mathbf{a}_t'$ to the trajectory optimizer to obtain a safe trajectory: $\mathbf{X} \leftarrow \text{TrajOpt}(\mathbf{s}_t, \mathbf{a}_t')$.
\State Initialize cumulative reward $r_t \leftarrow 0$. 
\State Initialize the initial state $\mathbf{s}_{t,i} \leftarrow \mathbf{s}_{t}$.
\For {$i=1$ to $k$}
\State Follow the trajectory:
\State $\mathbf{a}_{t,i} \leftarrow \text{TrajFollow}(\mathbf{s}_{t,i}, \mathbf{X},\pi_{\phi})$.
\State Execute $\mathbf{a}_{t,i}$ and observe reward $r_{t,i}$ and $\mathbf{s}_{t,i}$.
\State Sum the reward $r_t \leftarrow r_t + r_{t,i}$.
\EndFor
\State Save the final state: $\mathbf{s}_{t+1} \leftarrow \mathbf{s}_{t,k}$.
\State Store the transition $(\mathbf{s}_t, \mathbf{a}_t', \mathbf{s}_{t+1}, r_t)$ into $D$.
\State Update the policy $\pi_{\theta}$ on data from $D$.

\EndFor
\EndFor
\end{algorithmic}
\end{algorithm}

\subsection{Definition of ``Root'' Node}
\begin{figure}[htbp]
     \centering
     \begin{subfigure}[b]{0.22\textwidth}
         \centering
         \includegraphics[width=\textwidth]{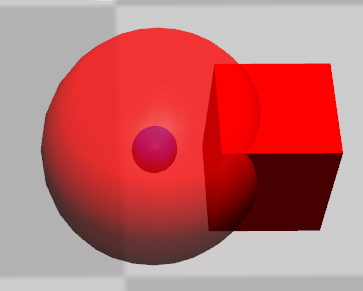}
         \caption{Point and Mass}
     \end{subfigure}
     \hfill
     \begin{subfigure}[b]{0.22\textwidth}
         \centering
         \includegraphics[width=\textwidth]{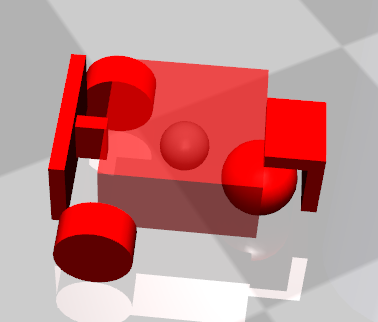}
         \caption{Car}
     \end{subfigure}
     \hfill
     \begin{subfigure}[b]{0.22\textwidth}
         \centering
         \includegraphics[width=\textwidth]{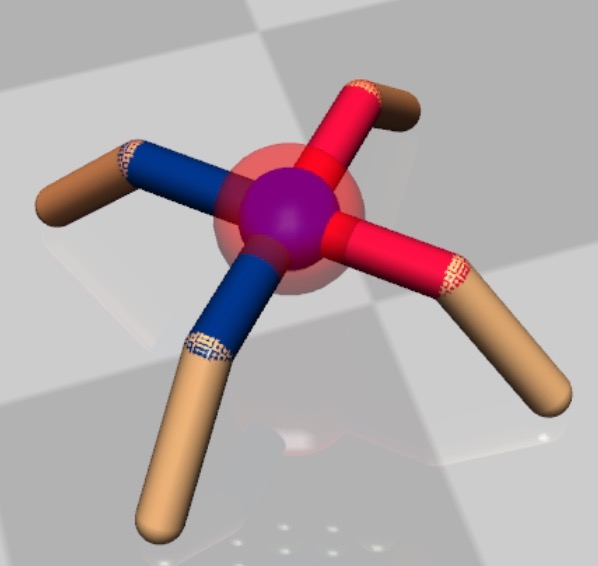}
         \caption{Ant}
     \end{subfigure}
        \caption{Root node position of different robots shown by the location of the small blue sphere.}
\label{fig:root_node}
\end{figure}

In MuJoCo, a robot is defined in a tree structure. Specifically, a robot is usually defined by mounting the adjacent link onto the previous link. The root node of the robot is the root body of the tree structure. In our experiments, it is usually the body of the robot. Thus the location of the root node can be interpreted as the center of the robot. The position of the root node is shown in Fig.~\ref{fig:root_node}. The small blue sphere inside the robot denotes the position of the root node.

\label{app_sec:root}
\subsection{Implementation Details about the RL policy}
We use Soft Actor-Critic (SAC)~\cite{haarnoja2018soft} to train the high-level policy. The actor network and the critic network are three-layer neural networks with a hidden size of 256. The learning rate for training is set to be $3\mathrm{e}{-4}$. The agent interacts with the environment for $1\mathrm{e}{7}$ time steps to ensure it has fully converged.
\subsection{Additional Details about the Trajectory-following Module}
\textbf{Training of the goal-following agent:} 
The goal $\mathbf{x}_i$ is sampled uniformly from a range of $d_{min}$ to $2 d_{min}$ away from the robot to match the input distribution during inference, in which $d_{min}$ is a distance threshold. The reward function used for training the goal-following agent is defined as the change in distance between the robot and the goal: 
\begin{equation}
    r^{g}_{t} := ||\mathbf{s}_{t-1, x} -\mathbf{g}|| - ||\mathbf{s}_{t,x} - \mathbf{g}||. 
\end{equation}
SAC is used to train the goal-following agent. The actor network and the critic network are three-layer neural networks with a hidden size of 256. $d_{min}$ is set to be $0.2m$. The agent interacts with the environment for $1\mathrm{e}{6}$ time steps to ensure it has fully converged.

\textbf{Choosing a waypoint to follow:} In the trajectory-following module, the trajectory is represented as a set of waypoints. However, the goal-following agent only takes in a single goal as input. Thus, we use the following procedure to select a waypoint from the set for the goal-following agent:
\begin{enumerate}
    \item The agent keeps track of the waypoint that is input into the goal-following agent (``tracking point"). The tracking point is initialized as the first waypoint of the trajectory. 
    \item At each time step, the agent searches from the tracking point to the end of the trajectory until it finds a waypoint that is at least $d_{min}$ away from the current robot root position. It sets the point as the tracking point for the current time step. It is possible that the tracking point is the same as the previous time step.
\end{enumerate}
\label{app_sec:traj_follow}
\subsection{Additional Details about the Trajectory Optimizer}
\label{app_sec::traj_opt_details}
\textbf{Hyperparameters: }
The distance threshold $\epsilon'$ is set to be $0.5m$ during training for all of the robots. The number of waypoints is set to be $30$. The maximum time step for optimizing the trajectory is set to be 10 steps during training. 
If the robot is currently within $\epsilon'$ of some obstacle, 
we would manually set $\boldsymbol{\lambda}$ to be the maximum value $\boldsymbol{\lambda}_{max}$ to encourage the robot to escape the unsafe region as quickly as possible. 

\textbf{Accelerate the training:}
In practice, since the Lagrangian method adds computational overhead, we use a fixed $\boldsymbol{\lambda}$ during training, which speeds up training time and also encourages exploration (although this leads to some safety violations during training). 
At test time, we update $\boldsymbol{\lambda}$  as mentioned previously, which ensures safety at test time.

\textbf{Implementation simplifications:}
In our experiments, instead of using three distinctive Lagrangian values for the three constraints mentioned in equation~\ref{equ:constraint_opt} respectively, we use a single Lagrangian value $\lambda$ for all 3 constraints, to simplify our implementation. 




To solve the dual problem mentioned in Equation~\ref{equ:lag}, instead of using gradient descent to update 
$\lambda$, in practice, we follow a simpler procedure of increasing $\lambda$ if the obstacle constraints are violated until the optimizer returns a feasible solution. This is effective since the cost function is always greater than or equal to 0. 
\subsection{Training Details in the Evaluation Experiment}
\label{app_sec:training_details}
For each experiment, we train four different seeds to account for randomness. 

For our method, we train it for $9\mathrm{e}{6}$ steps in all the environments. For CPO, PPO Lagrangian, TRPO Lagrangian, we train them for $6\mathrm{e}{7}$ in MassPush1, PointPush1, CarPush1, PointPush2, and CarPush2, and we train them for $1\mathrm{e}{7}$ in AntPush1. For Safety Editor and CVPO, we train them for $1\mathrm{e}{7}$ for all the environments. The baseline methods have more time steps to ensure convergence.

\subsection{Implementation Details of Training the Goal-reaching Policy}
For our method and the ``PPO Lag + SAC" ablation, we use SAC to train the goal-reaching policy. In our method, we sample the goal from a range of $d_{min}$ to $2d_{min}$. Specifically, $d_{min}$ is set to be $0.2m$ in our experiment to match the average distance between waypoints in the trajectory. For the ``PPO Lag + SAC" ablation, we sample the goal randomly from a $2m\times2m$ area to match the distribution of subgoals output from the high-level PPO Lagrangian  agent.

\subsection{Real-robot Experiment Details}
\label{app_sec:real_robot}
In the real-robot experiment, the fingertip of the gripper moves in a plane to push the box towards the goal (the large green circle). As in the simulation setup. the robot also needs to avoid hazards (the small red circle) and try not to get stuck by the pillar (blue).

In order to match the observation distribution seen during training, 
we convert the location of the objects into LiDAR readings in a similar format to the Safety Gym Pseudo LiDAR.

In this experiment, we first train the policy in simulation. We modify the simulation environment to match the setup in the real world, e.g., we change the size of objects and the shape of the robot in the simulation. Then we directly transfer the policy to the real robot without any finetuning.

\section{Safety Gym Details}
\label{app_sec:safety_gym}
\subsection{Safety Gym Objects}
The following types of objects exist in the environments for the Push tasks:
\begin{enumerate}
    \item \textbf{Goal} indicates where the robot needs to reach. As long as the robot is within a specific range of the goal, the task is considered successful.
    \item \textbf{Hazard} denotes circular regions that the robot should not enter. It is a virtual region and does not interact with the robot. The robot entering such a region will incur the cost. It is defined as obstacles in our method.
    \item \textbf{Pillar} is a fixed cylinder; contact with the pillar will not incur any cost. Though contacting with pillar might now incur cost, the robot might get stuck. Thus, it is also defined as an obstacle in our method.
    \item \textbf{Box} is an object that the robot can (and must) interact with: the goal is for the agent to push the box towards the goal. If the box enters the hazard regions, it will not incur any cost. 
\end{enumerate}
\subsection{Safety Gym agent}
There are four different kinds of agents in our Safety Gym experiments: Point, Car, Mass, and Ant.
The Point and Car are two default agents from the Safety Gym benchmark; the Point agent is a robot with one actuator for turning and another actuator for moving forward or backward. The Car agent is a robot with two independently driven wheels.  
We created the Mass agent, which is an omnidirectional agent; the action space of the Mass agent is defined as a delta movement of the agent position. The Ant agent is a quadrupedal agent with eight joints, similar to the Ant agent from MuJoCo~\cite{todorov2012mujoco}.

\subsection{Reward Function}
The reward function for the ``Push'' task is defined as the change in the distance between the robot and the box and between the box and the goal:
\begin{equation}
    r_{t}^{p} := ||\mathbf{x}^b_{t-1} - \mathbf{g}|| - ||\mathbf{x}^b_{t} - \mathbf{g}|| + ||\mathbf{s}_{t-1, x} - \mathbf{x}^b_{t-1}|| - ||\mathbf{s}_{t,x} - \mathbf{x}^b_{t}||,
\end{equation}
in which $\mathbf{x}^b_{t}$ denotes the location of the center of the box at the $t$~th step, and $\mathbf{g}$ denotes the location of the final goal that the box needs to reach. 
\subsection{Cost Function}
The cost function is defined as:  
\begin{equation}
    c_{t}^{p} := \mathbbm{1}(\min_j ||\mathbf{s}_{t,x} - \mathbf{x}_j^{obs}|| < \epsilon),
\end{equation}
in which $\mathbbm{1}$ is the indicator function, and $\epsilon$ is a distance threshold. The cost function will return $1$ if the ``root'' of the robot is within $\epsilon$ of any obstacle. 

\subsection{Observation Space}
The observation space of the Safety Gym environment consists of LiDAR observation and robot states. The LiDAR observation returns an approximate position of the objects, due to the limited LiDAR resolution. The robot states consist of robot velocity, acceleration, and orientation.


\section{Additional Analysis}
\subsection{Reasons of Safety Violations}
\label{app_sec:source_inaccuracy}
Though our trajectory optimizer is formulated to completely avoid obstacles, in practice, the executed trajectory might not be perfectly safe, due to perception errors or errors with the trajectory-following module. Specifically, there might be errors in practice due to a number of factors:
\begin{enumerate}
    \item Localization errors in estimating the robot position $\mathbf{s}_{t,x}$ 
    \item Perception errors in estimating the locations of the obstacles $\mathbf{x}_j^{obs}$
    \item Errors in following the planned trajectory
    \item Discretization errors: since the planned trajectory is a discrete set of waypoints $\mathbf{X} := \{\mathbf{x}_1, \mathbf{x}_2, ..., \mathbf{x}_N\}$, it is possible that the agent will encounter an unsafe state in between the trajectory waypoints that was not accounted for in the trajectory optimization.
    \item It is possible that no safe trajectory exists from the robot's current state.
\end{enumerate}
\subsection{Additional Limitations for SEMDP}
\textbf{The cost function is given:} The cost function needs to be known a priori for SEMDP, which is an additional strong assumption compared to the other safe RL and safe exploration methods. We argue that we focus on tasks which require avoiding obstacles. Obstacle avoidance is a very common safety criterion in robotics tasks and assuming knowing the requirement a prior will not lose much generality. 

\textbf{How can SEMDP generalize beyond navigation tasks:} We evaluate our algorithm on goal-reaching tasks and box-pushing tasks in our paper. Those tasks are essentially navigation tasks. Our algorithm framework is designed to generalize to any tasks that can be decomposed into a sequence of subgoals. The subgoal does not have to be a 3-D position where the robot root needs to reach. It can also be the end effector position that the robot needs to reach. Moreover, subgoals can also be defined as joint angles, which can be applied to manipulation tasks. The application of the framework in manipulation tasks will be left for future work. 

\subsection{Code Release}

Our code is available at \href{https://github.com/safetyembedded/SafetyEmbeddedMDP}{https://github.com/safetyembedded/SafetyEmbeddedMDP}.

\end{document}